\documentclass{INTERSPEECH2023}

\interspeechcameraready 
\usepackage{multirow}
\usepackage{amsmath}

\title{Few-Shot Open-Set Learning for On-Device \\ Customization of KeyWord Spotting Systems}
\name{Manuele Rusci$^1$, Tinne Tuytelaars$^1$}
\address{ 
  $^1$PSI, KU Leuven, Belgium}
\email{ \{manuele.rusci,tinne.tuytelaars\}@esat.kuleuven.be}

\begin{document}

\maketitle
 
\begin{abstract}
A personalized KeyWord Spotting (KWS) pipeline typically requires the training of a Deep Learning model on a large set of user-defined speech utterances, preventing fast customization directly applied on-device. 
To fill this gap, this paper investigates few-shot learning methods for open-set KWS classification by combining a deep feature encoder with a prototype-based classifier. With user-defined keywords from 10 classes of the Google Speech Command dataset, our study reports an accuracy of up to 76\% in a 10-shot scenario while the false acceptance rate of unknown data is kept to 5\%. 
In the analyzed settings, the usage of the triplet loss to train an encoder with normalized output features performs better than the prototypical networks jointly trained with a generator of dummy unknown-class prototypes. This design is also more effective than encoders trained on a classification problem and features fewer parameters than other iso-accuracy approaches. 
\end{abstract}


\section{Introduction}
Keyword Spotting, which is the ability to recognize speech commands or \textit{wake-words}, is getting popular among battery-powered smart audio sensors~\cite{lopez2021deep}.
Because a KWS detection pipeline is intended to run continuously on the target system, the reduction of computation and memory costs of Deep Learning based KWS algorithms have been extensively investigated over the last years~\cite{rybakov2020streaming,banbury2021micronets}. 
The design of a custom KWS algorithm typically demands the training of a model on a dataset of collected \textit{user-defined keywords}~\cite{zhang2017hello}. 
Despite its effectiveness, such a design process is subject to the availability of computing resources and speech recordings, preventing users from obtaining a custom solution in a short time, e.g., on-device.

Few-Shot Learning (FSL) provides a viable solution to deal with the scarcity of abundant user-defined keywords data. 
In the field of KWS, there are several recent FSL methods that rely on the Prototypical Network (ProtoNet) concept~\cite{snell2017prototypical}. 
During system setup in the target scenario, users are asked to provide a few enrollment samples for each keyword. 
These reference speech data are then processed through a trained feature encoder to produce a set of feature vectors. 
A \textit{class prototype} is then computed as the mean of the feature vectors for each user-defined keyword. 
When it comes to inference, the distances between the output feature vector of a test sample, or \textit{embedding}, and the class prototypes are calculated. 
The classification output is determined by the shortest distance.
%
%
To gain high accuracy, the ProtoNet's feature encoder is trained to cluster the embeddings of speech samples belonging to the same class. At the same time,  feature vectors of different classes are forced to be distant according to a given distance metric, e.g.~Euclidean.

With respect to FSL techniques involving fine-tuning 
on a few labeled data~\cite{mazumder21_interspeech}, the fit of a prototype-based classifier is a low-cost option that can be easily implemented on-device without the need for backpropagation.
At present, however, the ProtoNet-based FSL approaches have been primarily assessed for closed-set classification, i.e. the test categories match those of the training set~\cite{chen21u_interspeech,Parnami22,jung2022metric}. 
In contrast, we argue that a customized KWS method shall work in an open-set setting, to distinguish user-defined keywords from \textit{unknown} speech utterances. 
To this aim, the ProtoNet approach has been recently extended by jointly training a generator of dummy prototypes for the unknown class~\cite{kim22h_interspeech}. 
Unfortunately, this work makes use of training data sampled from the same distribution of the target data, i.e. different class subsets of the Google Speech Command dataset. 
In addition, other works showed angular variants of the prototypical loss to achieve the highest accuracy for closed-set FSL classification~\cite{jung2022metric} or efficiently learning the feature encoder using the triplet loss but not in a few-shot setting~\cite{vygon2021learning}.
Hence, we denoted the present literature on few-shot open-set learning KWS solutions to be highly fragmented and it is missing clear design guidelines for on-device KWS customization.

To bridge this gap, this paper contributes 
an evaluation framework for FSL architectures composed by a feature encoder and a prototype-based open-set classifier initialized with few-shot samples. 
More in detail, we leverage the recent Multilingual Spoken Words Corpus (MSWC) dataset~\cite{mazumder2021multilingual} to train a feature extractor using the prototypical loss, its angular variant or the triplet loss. 
The evaluation is performed on the Google Speech Command (GSC) dataset, which is partitioned between a collection of target keywords, i.e. the positive set, and a negative set of \textit{unknown} keywords. 
In our analysis we compare the open-set classifier featuring a \textit{dummy proto generator} with either a simple variant that computes the unknown-class prototype using few random words or an alternative based on OpenMAX~\cite{bendale2016towards}, which statistically models the distance of data samples from the class prototypes to estimate if a test sample can fit any of the known classes. 
When considering Depthwise-Separable Convolutional Neural Network (DSCNN) encoders tailored for low-power embedded systems~\cite{zhang2017hello}, we show that a training process using the triplet loss and normalized features brings a superior accuracy than a ProtoNet-based method for open-set FSL classification under a fixed training epoch budget.
Our code is available at: \url{https://github.com/mrusci/ondevice-fewshot-kws}.

\section{Related Work}
A set of works denoted as "Query-by-Example" investigated methods to personalize the wake-word of a KWS system after providing a reference utterance. 
\cite{Chen2015} trained a 2-layer LSTM and proposed to use a similarity score to compare the last embedding state with a reference vector. 
An LSTM encoder was also leveraged by~\cite{kim22h_interspeech} to output a hypothesis graph based on the phonetic-based posteriorgram for the comparison. This method is however expensive at inference time because the memory-bound LSTM workload occurs at any window step and may lead to a runtime 10-100$\times$ slower than our considered convolutional method applied on a large time window (\SI{1}{\second}).

Differently from~\cite{chen20j_interspeech} that proposed to use MAML to train an effective representation for FSL on KWS, 
authors of~\cite{mazumder21_interspeech} leveraged transfer learning on a large encoder model (EfficientNet) to train a classifier on the target scenario. 
Conversely, many recent works rely on the Prototypical Network approach~\cite{snell2017prototypical}, which does not require any training to encapsulate the information of the few-shot samples. 
Several studies focused only on closed-set problems and used the data from the same distribution (typically the GSC dataset) at training and test time~\cite{chen21u_interspeech,Parnami22}.
\textit{Jung et al.}~\cite{jung2022metric} showed the higher effectiveness of the angular prototypical loss on a KWS FSL scenario compared to classification-based encoders. 
After training a Resnet15 using data from 1000 classes of Librispeech, the authors proposed to fine-tune the feature extractor on the categories of the test dataset to gain an accuracy of up to 96\% for a 10-shot closed-set problem. 
A personalized KWS system~\cite{yang22l_interspeech} also leveraged the angular loss to train an additional model for keyword adaptation.
In parallel, ~\cite{vygon2021learning} proposed  the triplet loss to train a ResNet15-based feature extractor on LibriSpeech, which showed a top score of 97\% on GSC using a kNN with all the training data, i.e. not FSL. After, \cite{Huang21,huang22l_interspeech} exploited a soft-triplet loss, which combines triplet loss and softmax, 
 for a Query-by-Example  custom wake-word solution.
This class of works focused on FSL training methodologies but lacks clear design guidelines and evaluation in an open-set scenario for customized KWS solutions that we address.

In the open-set context, \cite{Huh21} used triplet or angular prototypical losses for KWS detection combined with a memory-expensive SVM classifier. However, this study do not consider few-shot evaluations and includes the user-defined keywords in the training dataset. 
More related to our study is the work of \textit{Kim et al.}~\cite{kim22h_interspeech} that we also reproduce in this paper. 
The authors used a ProtoNet setting and jointly trained a generator of dummy prototypes for the unknown class. 
Unfortunately, this study is limited by the usage of GSC for training and testing, and the evaluation is restricted to a 5-shot-5-way, in which this method scores up to 86.9\% in accuracy. 
PEELER~\cite{liu2020few} proposed to include an extra model to estimate the variance associated with every class prototype and rely on the Mahalanobis distance at inference time for open-set classification. 
This method is however memory-expensive because of the overhead to store the spatial variance tensors and the estimator.

\begin{figure*}[t]
  \centering
  \includegraphics[width=0.9\textwidth]{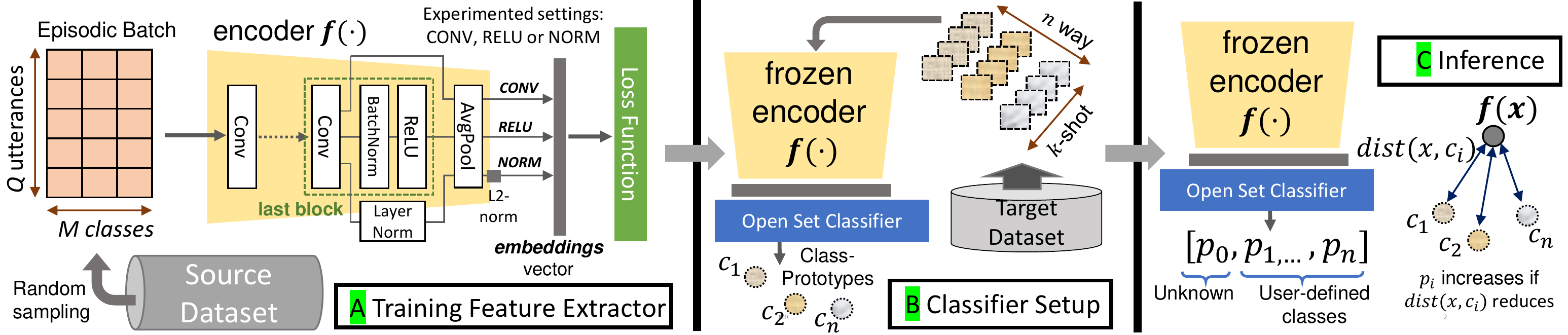}
  \caption{Overview of the approach considered for KWS open-set FSL. }
  \label{fig:method}
\end{figure*}

\section{Method}\label{sec:method}
Figure~\ref{fig:method} depicts the design flow considered for On-Device KWS customization.
The core algorithm is composed by a Deep Learning based feature encoder and an open-set classifier. 
Firstly, the feature encoder is trained offline (Fig.~\ref{fig:method}A), i.e. using a server machine before deployment on the target HW, on a large source dataset of labelled spoken keywords. 
At test time, an open-set classifier is plugged on top of the feature extractor and initialized with few-shot utterances taken from the target scenario (Fig.~\ref{fig:method}B).
We consider the target dataset to be disjoined from the training data; test utterances belong to classes not represented in the source dataset. Train and test data are also sampled from different distributions. 
As shown in Fig.~\ref{fig:method}C, the inference is based on a distance score accounting for the unknown class as explained in Sec.~\ref{sec:classifier}.

\subsection{Training the Feature Encoder}\label{sec:learning}

The feature encoder $f(\cdot)$ is trained in a supervised fashion following an episode-based protocol. 
At every episode, the dataloader feeds the model with a batch of training data fetched from the source dataset.
Every batch features a balanced number of samples from $M$ randomly chosen classes. 
In the following, we detail the considered training strategies.

\textbf{Prototypical Network (PN)}. 
Let us consider an episodic batch that includes S support samples $\{x^S_{i,j}\}_{i=1}^S$ and Q query samples $\{x^Q_{i,j}\}_{i=1}^Q$ for every class $j=1,..,M$, for a total of $(S+Q)\times M$ utterances per batch. 
At every episode, a set of prototypes $c = \{c_j\}^\mathrm{M}_{j=1}$ is firstly computed as: 
\begin{equation}\label{eq:proto}
    c_j = \frac{1}{S} \sum_{i=1}^S {f(x^S_{i,j})}
\end{equation}
Then, the loss function minimizes the negative log probability of correctly predicting the category of the query samples:
\begin{equation}\label{eq:PN}
  L_{PN} = - \frac{1}{Q \cdot M} \sum_{i=1}^Q \sum_{j=1}^M  \mathrm{log} \frac{\mathrm{exp}(\textbf{s}(x^Q_{i,j}, j))}{\sum^M_{k=1} \mathrm{exp}( \textbf{s}(x^Q_{i,j}, k) )}
\end{equation}

where 
\begin{equation}\label{eq:distL2}
  \textbf{s}(x, j) = -d_{L2}(f(x), c_j)
\end{equation}
and  $d_{L2}$ is the euclidean distance. 

\textbf{Angular Prototypical (AP)}. Following~\cite{chung20b_interspeech,jung2022metric}, we replace the euclidean distance in Eq.~\ref{eq:distL2}  with the cosine similarity within the same formulation of the Prototypical Network (Eq.~\ref{eq:PN}) as: 

\begin{equation}\label{eq:AL}
  \textbf{s}(x, j) = w \cdot (\mathrm{cos}(f(x), c_j)-m) + b
\end{equation}
where $w$ and $b$ are learnable scalar parameters, $c_j$ is computed from Eq.~\ref{eq:proto} and $m$ is a margin different from zero for the score of the true class. 

\textbf{Triplet Loss (TL)}. 
The triplet loss is computed based on $N_{t}$ triplets $\{x_{i},x^+_{i},x^-_{i} \}_{i=1}^{N_{t}}$ sampled from an episodic batch:

\begin{equation}\label{eq:TL}
\resizebox{0.91\hsize}{!}{$L_{TL} = \frac{1}{N_{t}} \sum_{i=1}^{N_{t}}  \mathrm{max}(0, d_{L2}(x_{i},x^+_{i}) - d_{L2}(x_{i},x^-_{i}) + m ) $}
\end{equation}
where $x^+_{i}$ belongs to the same category of $x_{i}$ and $x^-_{i}$ is selected from a different random class. $m$ is the margin. 

\subsubsection{Feature Encoder Architecture}
Without loss of generality, this study considers the DSCNN model as the feature encoder, which is largely used for on-device KWS~\cite{zhang2017hello}. 
This deep model is composed of a stacked sequence of depthwise and pointwise convolution blocks.
Every block includes a BatchNorm and a ReLU activation after the convolution. 
We experiment by passing to the final average pooling layer the feature maps produced by the last convolution or the ReLU layers. 
These feature extractor versions are denoted respectively as \textit{DSCNN-CONV} or \textit{DSCNN-RELU} (see Fig.~\ref{fig:method}A).
Note that the latter includes only non-negative values in the high-dimensional space.
In addition, we also consider the case of normalized output features \cite{vygon2021learning}. Following \cite{zhai2018classification}, the \textit{DSCNN-NORM} replaces the last BatchNorm layer with a LayerNorm and applies L2 normalization after the average pooling. 
In this last setting, the feature encoder is forced to discriminate classes based on the phase of the embeddings.

\subsection{Open-Set Classification}\label{sec:classifier}
At inference time, the KWS pipeline receives $\mathrm{K}$ enrollment samples for every \textit{user-defined keyword}.
We consider $\mathrm{N}$ new keywords in the target domain and, therefore, 
a $\mathrm{K}$-shot $\mathrm{N}$-way classification problem.
After computing the embeddings of the $\mathrm{K}$-shot samples using the trained encoder, a classifier computes (Eq. \ref{eq:proto}) and stores the class prototypes. 

When a new test sample is fed to the pipeline, the open-set classifier returns a probability score vector $P=\{p_{i}\}_{i=0}^N$ based on the current prototype set. 
Specifically, $p_0$ is the prediction score for the \textit{unknown class} and $p_{i\neq0}$ is the probability of the i-th keyword, which assumes the highest value if the  distance score of Eq.~\ref{eq:distL2} is the lowest.  The final class prediction $y$ is:

\begin{equation}\label{eq:classifier}
    y = 
    \begin{cases}
    \mathrm{arg}\  \mathrm{max}\quad  p_i,& \text{if } p_i \geq \gamma\\
    0,              & \text{otherwise}
\end{cases}
\end{equation}
where $y=0$ denotes the unknown class and $\gamma$ is a manually-tunable parameter to tradeoff between the classifier's precision and recall. 
In the following, we detail the considered variants for the open-set classifiers.

\textbf{Open Nearest Class Mean (openNCM)}~\cite{hayes2022online}. 
The Nearest Class Mean classifier is typically adopted by Prototypical Networks. 
A simple open-set variant estimates the $c_0$ prototype for the \textit{unknown} class using K random samples taken from the target domain but not belonging to the user classes. 
The probability score is computed by applying the SoftMax on the (N+1)-sized distance vector obtained from Eq.~\ref{eq:distL2}. 
If the feature encoder is trained using the AP loss, the classifier applies L2 normalization before computing the euclidean distance.

\textbf{OpenMAX}~\cite{bendale2016towards}. 
Following~\cite{mundt2023wholistic}, for every known class the distance from the prototype vector is statistically modeled using a Weibull distribution (we feed the 5 largest distances from the K enrollment samples to the \texttt{fit\_high} function). 
At test time, the distances of a test sample from the known classes (Eq.~\ref{eq:distL2}) are scaled by a factor $w^{wb}_i \in [0,1]$, which tends to 1 if the current sample is on the tail of the \textit{i}-th Weibull class distribution. The score for the unknown category is then estimated with Eq.~\ref{eq:openmax}  before applying the Softmax function. 

\begin{equation}\label{eq:openmax}
  \textbf{s}(x, j=0) = - \sum_{i=1}^N (1- w^{wb}_i)\cdot d_{L2}(f(x), c_i)
\end{equation}

\textbf{Dummy Proto (DProto)}~\cite{kim22h_interspeech}. 
A generator $g(\cdot)$ of dummy prototypes for the unknown category is jointly trained with a ProteNet-based feature encoder $f(\cdot)$. 
Thanks to this, the unknown prototype is estimated at inference time as $c_0 = g(\{c_j\}^\mathrm{N}_{j=1})$. 
Driven by the results of the original paper~\cite{kim22h_interspeech}, $g(\cdot)$ is trained to produce 3 unknown prototypes\footnote{We rely on our implementation because not any code is available.}; only the closest to the test samples is selected at inference time to compute the final probability score, in a similar way to openNCM. 
At training time, data samples from a subset of categories in the episodic batch are assigned to the unknown class to jointly learn the feature extractor and the generator $g$. 
We remind the reader to the original paper for more details.

\subsection{Evaluation Framework} \label{sec:eval_framework}
We use the \textit{English} partition of the Multilingual Spoken Words Corpus (MSWC) dataset~\cite{mazumder2021multilingual} to train the feature encoder. 
We restrict the data sampling to the 500 classes with the highest number of utterances after excluding the categories of the test dataset (the $\mathrm{GSC+}$ partition described below).
The train samples are augmented with additive background noises taken from the DEMAND dataset~\cite{thiemann2013diverse}. 
Noise is applied according to a uniform probability of 0.95 and a random SNR between 0 and 5.

For the open-set testing, we refer to the Google Speech Command (GSC) dataset~\cite{warden2018speech}. 
The GSC dataset is composed of {\SI{1}\second}  long speech utterances from 35 categories. 
For our experiments, we define a positive ($\mathrm{GSC+}$) and a negative ($\mathrm{GSC-}$) partition. 
The GSC+ includes only samples belonging to 10 classes (\textit{yes}, \textit{no}, \textit{up}, \textit{down}, \textit{left}, \textit{right}, \textit{on}, \textit{off}, \textit{stop}, \textit{go}), which constitute the target categories for few-shot classification.
The K enrollments to setup the classifier within a K-shot N-way test are sampled from the original \textit{train} split of the GSC dataset. 
In the case of the openNCM, the prototype of the \textit{unknown} class is computed based on data randomly fetched from 5 classes:  \textit{backward}, \textit{forward}, \textit{visual}, \textit{follow}, \textit{learn}.
The $\mathrm{GSC-}$ partition includes the test samples from the remaining 20 classes. 

To assess the performance of every tested configuration we report: 
\textit{(i)}  the classification accuracy on the $\mathrm{GSC+}$ partition when $\gamma$ is tuned to achieve a False Acceptance Rate (FAR)\footnote{Ratio of negative samples classified as positives} of 5\% on the $\mathrm{GSC-}$ dataset ($ACC^+_{5\%}$),
\textit{(ii)}  the False Rejection Rate at FAR of 5\% ($FRR^+_{5\%}$),
and \textit{(ii)} the $AUROC$. 
The classification metrics are measured over the data samples fetched from the \textit{test} split of GSC 
and we provide the mean statistics over 10 test repetitions. 

\begin{table*}[t]
    \centering
    \caption{Performance on the GSC testset under \{5,10\}-shot 10-way open-set classification settings.}
    \label{tab:big}
    \resizebox{\textwidth}{!}{
        \begin{tabular}{|cc|ccc|ccc|ccc|ccc|ccc|ccc|}
\toprule
 \textbf{Loss} &  \textbf{Feature} & \multicolumn{3}{c}{\textbf{openNCM  5-shot}} & \multicolumn{3}{c}{\textbf{OpenMAX 5-shot}} & \multicolumn{3}{c|}{\textbf{Dproto 5-shot}} & \multicolumn{3}{c}{\textbf{openNCM   10-shot}} & \multicolumn{3}{c}{\textbf{OpenMAX 10-shot}} & \multicolumn{3}{c|}{\textbf{Dproto 10-shot}} \\
 & \textbf{Extractor} & \multicolumn{1}{l}{\scriptsize $ACC^+_{5\%}$} & \multicolumn{1}{l}{\scriptsize AUROC} & \multicolumn{1}{l|}{\scriptsize $FRR^+_{5\%}$} & \multicolumn{1}{l}{\scriptsize $ACC^+_{5\%}$} & \multicolumn{1}{l}{\scriptsize AUROC} & \multicolumn{1}{l|}{\scriptsize $FRR^+_{5\%}$} & \multicolumn{1}{l}{\scriptsize $ACC^+_{5\%}$} & \multicolumn{1}{l}{\scriptsize AUROC} & \multicolumn{1}{l|}{\scriptsize $FRR^+_{5\%}$} & \multicolumn{1}{l}{\scriptsize $ACC^+_{5\%}$} & \multicolumn{1}{l}{\scriptsize AUROC} & \multicolumn{1}{l|}{\scriptsize $FRR^+_{5\%}$} & \multicolumn{1}{l}{\scriptsize $ACC^+_{5\%}$} & \multicolumn{1}{l}{\scriptsize AUROC} & \multicolumn{1}{l|}{\scriptsize $FRR^+_{5\%}$} & \multicolumn{1}{l}{\scriptsize $ACC^+_{5\%}$} & \multicolumn{1}{l}{\scriptsize AUROC} & \multicolumn{1}{l|}{\scriptsize $FRR^+_{5\%}$} \\
 \midrule
\multirow{3}{*}{PN} & 
DSCNN-L-NORM & 
0.21 & 0.66 & 0.78 & 
0.23 & 0.79 & 0.77 & 
0.21 & 0.64 & 0.79 & 
0.22 & 0.68 & 0.78 & 
0.23 & 0.75 & 0.77 & 
0.22 & 0.67 & 0.78 
\\
 & DSCNN-L-CONV & 
 0.54 & 0.86 & 0.46 & 
 0.12 & 0.87 & 0.87 & 
 0.64 & 0.91 & 0.35 & 
 0.62 & 0.89 & 0.37 & 
 0.48 & 0.89 & 0.50 & 
 0.71 & 0.93 & 0.28 
 \\
 & DSCNN-L-RELU & 
 \textbf{0.56 } &  \textbf{0.87  }&  \textbf{0.43 }& 
 0.14 & 0.91 & 0.85 & 
  \textbf{0.66 }& \textbf{ 0.92} &  \textbf{0.32} & 
  \textbf{0.63} &  \textbf{0.89} &  \textbf{0.37} & 
 \textbf{0.56} & \textbf{0.92} & \textbf{0.40} & 
  \textbf{0.71 }&  \textbf{0.93} &  \textbf{0.28} 
 \\
 \hline

AP & DSCNN-L-NORM & 
\textbf{0.66 }& \textbf{0.92} & \textbf{0.29 }& 
0.44 & 0.94 & 0.54 & 
 \textbf{0.65} &  \textbf{0.93} & \textbf{0.30 }& 
\textbf{0.71 }& \textbf{0.93 }& \textbf{0.25 }& 
\textbf{0.66 }& \textbf{0.93 }& \textbf{0.30 }& 
\textbf{0.70 }& \textbf{0.94 }& \textbf{0.25 }
\\
\hline
\multirow{3}{*}{TL} & 
DSCNN-L-NORM & 
\textbf{0.71} & \textbf{0.93} & \textbf{0.26} & 
0.37 & 0.94 & 0.62 & 
&  &  & 
\textbf{0.76 }& \textbf{0.94} &\textbf{ 0.21} & 
\textbf{0.71 }& \textbf{0.94}& \textbf{0.24} &  
&  &  
\\

 & DSCNN-L-CONV & 
 0.58 & 0.88 & 0.41 &
 0.25 & 0.95 & 0.74 &  
 &  &  & 
 0.63 & 0.89 & 0.36 & 
 0.67 & 0.95 & 0.29 &  
 &  &  \\
 & DSCNN-L-RELU & 
 0.66 & 0.90 & 0.33 & 
 0.20 & 0.96 & 0.80 & 
 &  &  & 
 0.71 & 0.91 & 0.28 & 
 0.64 & 0.96 & 0.32 & 
 &  &  \\
  \midrule

\multirow{3}{*}{PN} & 
DSCNN-S-NORM & 
0.14 & 0.57 & 0.85 & 
0.14 & 0.72 & 0.85 & 
0.14 & 0.54 & 0.85 & 
0.17 & 0.59 & 0.83 & 
0.17 & 0.69 & 0.83 & 
0.15 & 0.55 & 0.84 \\
 & DSCNN-S-CONV & 
 \textbf{0.40} &  \textbf{0.81} &  \textbf{0.60} &
 0.14 & 0.83 & 0.84 & 
 \textbf{0.40} &  \textbf{0.81} &  \textbf{0.59} & 
 \textbf{0.48} &  \textbf{0.85} &  \textbf{0.51} & 
 0.38 & 0.86 & 0.60 & 
 0.43 & 0.83 & 0.57 \\
 & DSCNN-S-RELU & 
 0.39 & {0.80} & {0.60} & 
 0.20 & 0.86 & 0.77 & 
 {0.39} & {0.80} & {0.60} & 
 {0.45} & {0.84} & {0.54} & 
 \textbf{0.44 }& \textbf{0.87 }& \textbf{0.54 }& 
 \textbf{0.44} &\textbf{ 0.81} & \textbf{0.56} \\ 
 \hline
AP & DSCNN-S-NORM & 
\textbf{0.39} & \textbf{0.83} & \textbf{0.60} & 
0.34 & 0.87 & 0.64 & 
0.31 & 0.81 & 0.68 & 
\textbf{0.41} & \textbf{0.84} & \textbf{0.57} & 
\textbf{0.36} & \textbf{0.86 }& \textbf{0.63} & 
0.33 & 0.82 & 0.66 \\
 \hline
\multirow{3}{*}{TL} & 
DSCNN-S-NORM & 
\textbf{0.51} & \textbf{0.87} & \textbf{0.46} & 
0.38 & 0.91 & 0.59 &  
&  &  & 
\textbf{0.56} & \textbf{0.89} & \textbf{0.42} & 
0.54 & 0.91 & 0.42 & 
&  &  \\
 & DSCNN-S-CONV & 
 0.39 & 0.80 & 0.60 & 
 0.26 & 0.92 & 0.70 & 
 &  &  & 
 0.42 & 0.82 & 0.57 & 
 0.56 & 0.92 & 0.39 & 
 &  &  \\
 & DSCNN-S-RELU & 
 0.42 & 0.82 & 0.57 & 
 0.28 & 0.92 & 0.69 & 
 &  &  & 
 0.49 & 0.85 & 0.50 & 
 \textbf{0.58} & \textbf{0.93} & \textbf{0.37} & 
 &  & \\
 \bottomrule

\end{tabular}
    }
\end{table*}

\section{Experimental Result}
In this study we consider the Large and Small versions of the DSCNN, respectively indicated as DSCNN-L and DSCNN-S and featuring 407k and 22k parameters~\cite{zhang2017hello}. 
The DSCNN model is fed with a MFCC feature map computed with a window size of \SI{40}{\milli\second} and a stride of 50\%. We rely on power spectrograms to avoid the costly square root operations and select the first 10 MFCC features to obtain 49x10 feature maps, as done in~\cite{zhang2017hello}. 
The size of the embedding vectors is 64 and 256 for DSCNN-S and DSCNN-L, repsectively.

Feature extractors are trained for a fixed duration of 40 epochs of 400 episodes. 
Adam optimizer is adopted with an initial learning rate of 0.001 and decayed by 0.1 after 20 epochs.
The Triplet Loss is computed over episodic batches of 20 samples $\times$ 80 classes; triplet negatives are randomly selected among the batch data of a different category. A margin $m$ of 0.5 is used. 
In the case of PN or AP, the dataloader samples 10 support samples and 30 query samples from 40 classes at every episode. 
For AP, $m$  is set to 0.5. 
To train the DProto feature extractor jointly with the dummy prototype generator, 16 of the 40 classes are marked as unknown.

\begin{table}[b]
    \centering
    \caption{10-shot 10-way open-set Classification Comparison.}
    \label{tab:comp}
    \resizebox{\linewidth}{!}{
          \begin{tabular}{ccccc }

\toprule 
\textbf{DSCNN-L} | params: 407k & $ACC^+_{5\%}$ & AUROC & Train Data & Extra Params
\\
	 \midrule
			
openNCM+\textit{Classif}~\cite{zhai2018classification}+ NORM	& 0.52 & 0.89& \textit{source} & - \\
openNCM+TL+NORM	&	0.76	&	0.94	&	\textit{source}	&	- \\
dProto~\cite{kim22h_interspeech}+RELU	&	0.71	&	0.93	&	\textit{source}	&	- \\
PEELER~\cite{liu2020few}	&	0.76	&	0.94 &	\textit{source}	&	+6.3M \\

\textit{end-to-end}~\cite{zhang2017hello}	& 0.76 & 0.93 &	\textit{target} & - \\ 
\bottomrule
\toprule

\textbf{DSCNN-S} | params: 22k & $ACC^+_{5\%}$ & AUROC & Train Data & Extra Params
\\
\midrule
				
openNCM+\textit{Classif}~\cite{zhai2018classification}+ NORM	&	0.47	&0.85	&	\textit{source}	&	- \\
openNCM+TL+NORM	&	0.56	&	0.89	&	\textit{source}	&	-\\
dProto~\cite{kim22h_interspeech}+RELU		&	0.44	& 0.82	&	\textit{source}	&	- \\
PEELER~\cite{liu2020few}		&	0.60	&	0.88	&	\textit{source}	&	+341k \\

\textit{end-to-end}~\cite{zhang2017hello}		&	0.72 & 0.93 & \textit{target} & - \\ 

\bottomrule

  \end{tabular}
    }
\end{table}

Table~\ref{tab:big} includes the results of the experiments, which are averaged over 3 runs to account uncertainty from different training seeds.
%
Firstly, we notice the PN loss to achieve the best results when a large DSCNN-RELU is used. 
This configuration is slightly better than DSCNN-CONV while the NORM counterpart is leading to a low score, denoting a low-class separability. 
A $<$2\% accuracy gap between the RELU and CONV configurations is also observed for the small DSCNN, with the latter scoring the best. 
More in detail, openNCM achieves the highest accuracies of 56\% and 63\%  for, respectively, 5- and 10-shot 10-way KWS classification.
In the 5-shot scenario, OpenMAX leads to a low score, meaning a correct statistical model is not inferred from few samples, as also demonstrated from all other analyzed cases. 
Conversely, the accuracy of OpenMAX is close to openNCM with 10 samples per-class. 
These observations are preserved when the model is trained using the DProto approach, achieving  up to +8\% accuracy and -9\% FRR compared to openNCM for DSCNN-L in 10-shot configuration.

Overall, the triplet loss achieves the highest accuracies of 71\% and 76\% with openNCM and DSCNN-L-NORM for 5-shot and 10-shot, respectively. 
Differently from PN, the NORM feature extractor leads to superior performance than CONV and RELU in the case of DSCNN-L with a 4-5\% accuracy improvement. 
We hypothesize the NORM feature extractor to foster the class separation process based on the phase of the embeddings. 
The performance difference between NORM and RELU is reduced on the small model up to inverting the trend when a 10-shot OpenMAX classifier is plugged on the DSCNN-S-RELU (+4\%). 
%
On the other side, the AP loss achieves a slightly lower accuracy than the TL but still gains 8-10\% accuracy with respect to the PN loss on DSCNN-L, either with openNCM or DProto classifiers. 
In the case of DSCNN-S, AP leads to a similar accuracy level of the PN losses.

Lastly, Table~\ref{tab:comp} shows the performance of the best configurations of TL+openNCM and Dproto on a 10-shot open-set problem along with other methods.
More in detail, we consider a DSCNN feature extractor trained on a classification problem following~\cite{zhai2018classification}  and  PEELER~\cite{liu2020few}.
Both methods are trained on MSWC and tested on GSC, according to our evaluation framework.
The usage of the TL leads to the same accuracy level as PEELER on DSCNN-L but without paying the overhead of the additional network that estimates the class variances ($+6.3\mathrm{M}$ parameters), while PEELER scores +5\% accuracy on DSCNN-S.  
On the other side, the classifier-based feature extractor performs poorly in open-set because it is overconfident in the case of wrong predictions. 
For comparison purposes, we also trained the DSCNN models end-to-end on GSC+ with an \textit{unknown} class composed by the 5 keywords listed in Sec.~\ref{sec:eval_framework}. 
Also in this case, we observe that a TL+openNCM achieves a similar $ACC^+_{5\%}$ in open-set.
Despite the high accuracy on known classes, i.e. up to 95\% if $\gamma=0$, the end-to-end classifier fails to predict unknown classes not seen during training, i.e. low accuracy on the negative set. 
Conversely, the lower model capacity of DSCNN-S brings the end-to-end approach to perform best if trained on the full target dataset.

\section{Conclusion}
This paper investigated few-shot open-set methods to again on-device KWS customization. 
We built a framework to evaluate multiple open-set classifiers placed on top of feature extraction modules and initialized with few enrollments.
Among the analyzed feature extractor variants trained on the MSWC dataset, we showed the triplet loss applied on normalized output features of a DSCNN-L model leads to the highest accuracy when coupled with an openNCM classifier, surpassing solutions based on prototypical networks. 
In the considered scenario, this solution is also comparable with the more complex PEELER method or an end-to-end model trained on the target dataset. 

\section{Acknowledgements}
This work is partly supported by the European Horizon Europe program under grant agreement 101067475.

\newpage

\bibliographystyle{IEEEtran}
\bibliography{mybib}

\begin{thebibliography}{10}
\providecommand{\url}[1]{#1}
\csname url@samestyle\endcsname
\providecommand{\newblock}{\relax}
\providecommand{\bibinfo}[2]{#2}
\providecommand{\BIBentrySTDinterwordspacing}{\spaceskip=0pt\relax}
\providecommand{\BIBentryALTinterwordstretchfactor}{4}
\providecommand{\BIBentryALTinterwordspacing}{\spaceskip=\fontdimen2\font plus
\BIBentryALTinterwordstretchfactor\fontdimen3\font minus
  \fontdimen4\font\relax}
\providecommand{\BIBforeignlanguage}[2]{{%
\expandafter\ifx\csname l@#1\endcsname\relax
\typeout{** WARNING: IEEEtran.bst: No hyphenation pattern has been}%
\typeout{** loaded for the language `#1'. Using the pattern for}%
\typeout{** the default language instead.}%
\else
\language=\csname l@#1\endcsname
\fi
#2}}
\providecommand{\BIBdecl}{\relax}
\BIBdecl

\bibitem{lopez2021deep}
I.~L{\'o}pez-Espejo, Z.-H. Tan, J.~H. Hansen, and J.~Jensen, ``Deep spoken
  keyword spotting: An overview,'' \emph{IEEE Access}, vol.~10, pp. 4169--4199,
  2021.

\bibitem{rybakov2020streaming}
O.~Rybakov, N.~Kononenko, N.~Subrahmanya, M.~Visontai, and S.~Laurenzo,
  ``Streaming keyword spotting on mobile devices,'' \emph{Proc. Interspeech},
  pp. 2277--2281, 2020.

\bibitem{banbury2021micronets}
C.~Banbury, C.~Zhou, I.~Fedorov, R.~Matas, U.~Thakker, D.~Gope,
  V.~Janapa~Reddi, M.~Mattina, and P.~Whatmough, ``Micronets: Neural network
  architectures for deploying tinyml applications on commodity
  microcontrollers,'' \emph{Proceedings of Machine Learning and Systems},
  vol.~3, pp. 517--532, 2021.

\bibitem{zhang2017hello}
Y.~Zhang, N.~Suda, L.~Lai, and V.~Chandra, ``Hello edge: Keyword spotting on
  microcontrollers,'' \emph{arXiv preprint arXiv:1711.07128}, 2017.

\bibitem{snell2017prototypical}
J.~Snell, K.~Swersky, and R.~Zemel, ``Prototypical networks for few-shot
  learning,'' \emph{Advances in neural information processing systems},
  vol.~30, 2017.

\bibitem{mazumder21_interspeech}
M.~Mazumder, C.~Banbury, J.~Meyer, P.~Warden, and V.~J. Reddi, ``{Few-Shot
  Keyword Spotting in Any Language},'' in \emph{Proc. Interspeech}, 2021, pp.
  4214--4218.

\bibitem{chen21u_interspeech}
Y.~Chen, T.~Ko, and J.~Wang, ``{A Meta-Learning Approach for User-Defined
  Spoken Term Classification with Varying Classes and Examples},'' in
  \emph{Proc. Interspeech}, 2021, pp. 4224--4228.

\bibitem{Parnami22}
\BIBentryALTinterwordspacing
A.~Parnami and M.~Lee, ``Few-shot keyword spotting with prototypical
  networks,'' in \emph{2022 7th International Conference on Machine Learning
  Technologies (ICMLT)}.\hskip 1em plus 0.5em minus 0.4em\relax New York, NY,
  USA: Association for Computing Machinery, 2022, p. 277–283. [Online].
  Available: \url{https://doi.org/10.1145/3529399.3529443}
\BIBentrySTDinterwordspacing

\bibitem{jung2022metric}
J.~Jung, Y.~Kim, J.~Park, Y.~Lim, B.-Y. Kim, Y.~Jang, and J.~S. Chung, ``Metric
  learning for user-defined keyword spotting,'' in \emph{IEEE International
  Conference on Acoustics, Speech and Signal Processing (ICASSP)}.\hskip 1em
  plus 0.5em minus 0.4em\relax IEEE, 2023, pp. 1--5.

\bibitem{kim22h_interspeech}
B.~Kim, S.~Yang, I.~Chung, and S.~Chang, ``{Dummy Prototypical Networks for
  Few-Shot Open-Set Keyword Spotting},'' in \emph{Proc. Interspeech}, 2022, pp.
  4621--4625.

\bibitem{vygon2021learning}
R.~Vygon and N.~Mikhaylovskiy, ``Learning efficient representations for keyword
  spotting with triplet loss,'' in \emph{Speech and Computer: 23rd
  International Conference, SPECOM 2021, St. Petersburg, Russia, September
  27--30, 2021, Proceedings 23}.\hskip 1em plus 0.5em minus 0.4em\relax
  Springer, 2021, pp. 773--785.

\bibitem{mazumder2021multilingual}
M.~Mazumder, S.~Chitlangia, C.~Banbury, Y.~Kang, J.~M. Ciro, K.~Achorn,
  D.~Galvez, M.~Sabini, P.~Mattson, D.~Kanter \emph{et~al.}, ``Multilingual
  spoken words corpus,'' in \emph{Thirty-fifth Conference on Neural Information
  Processing Systems Datasets and Benchmarks Track (Round 2)}, 2021.

\bibitem{bendale2016towards}
A.~Bendale and T.~E. Boult, ``Towards open set deep networks,'' in
  \emph{Proceedings of the IEEE conference on computer vision and pattern
  recognition}, 2016, pp. 1563--1572.

\bibitem{Chen2015}
G.~Chen, C.~Parada, and T.~N. Sainath, ``Query-by-example keyword spotting
  using long short-term memory networks,'' in \emph{IEEE International
  Conference on Acoustics, Speech and Signal Processing (ICASSP)}, 2015, pp.
  5236--5240.

\bibitem{chen20j_interspeech}
Y.~Chen, T.~Ko, L.~Shang, X.~Chen, X.~Jiang, and Q.~Li, ``{An Investigation of
  Few-Shot Learning in Spoken Term Classification},'' in \emph{Proc.
  Interspeech}, 2020, pp. 2582--2586.

\bibitem{yang22l_interspeech}
S.~Yang, B.~Kim, I.~Chung, and S.~Chang, ``{Personalized Keyword Spotting
  through Multi-task Learning},'' in \emph{Proc. Interspeech}, 2022, pp.
  1881--1885.

\bibitem{Huang21}
J.~Huang, W.~Gharbieh, H.~S. Shim, and E.~Kim, ``Query-by-example keyword
  spotting system using multi-head attention and soft-triple loss,'' in
  \emph{IEEE International Conference on Acoustics, Speech and Signal
  Processing (ICASSP)}, 2021, pp. 6858--6862.

\bibitem{huang22l_interspeech}
J.~Huang, W.~Gharbieh, Q.~Wan, H.~S. Shim, and H.~C. Lee, ``{QbyE-MLPMixer:
  Query-by-Example Open-Vocabulary Keyword Spotting using MLPMixer},'' in
  \emph{Proc. Interspeech}, 2022, pp. 5200--5204.

\bibitem{Huh21}
J.~Huh, M.~Lee, H.~Heo, S.~Mun, and J.~S. Chung, ``Metric learning for keyword
  spotting,'' in \emph{2021 IEEE Spoken Language Technology Workshop (SLT)},
  2021, pp. 133--140.

\bibitem{liu2020few}
B.~Liu, H.~Kang, H.~Li, G.~Hua, and N.~Vasconcelos, ``Few-shot open-set
  recognition using meta-learning,'' in \emph{Proceedings of the IEEE/CVF
  Conference on Computer Vision and Pattern Recognition}, 2020, pp. 8798--8807.

\bibitem{chung20b_interspeech}
J.~S. Chung, J.~Huh, S.~Mun, M.~Lee, H.-S. Heo, S.~Choe, C.~Ham, S.~Jung, B.-J.
  Lee, and I.~Han, ``{In Defence of Metric Learning for Speaker Recognition},''
  in \emph{Proc. Interspeech}, 2020, pp. 2977--2981.

\bibitem{zhai2018classification}
A.~Zhai and H.-Y. Wu, ``Classification is a strong baseline for deep metric
  learning,'' \emph{arXiv preprint arXiv:1811.12649}, 2018.

\bibitem{hayes2022online}
T.~L. Hayes and C.~Kanan, ``Online continual learning for embedded devices,''
  in \emph{Conference on Lifelong Learning Agents}.\hskip 1em plus 0.5em minus
  0.4em\relax PMLR, 2022, pp. 744--766.

\bibitem{mundt2023wholistic}
M.~Mundt, Y.~Hong, I.~Pliushch, and V.~Ramesh, ``A wholistic view of continual
  learning with deep neural networks: Forgotten lessons and the bridge to
  active and open world learning,'' \emph{Neural Networks}, 2023.

\bibitem{thiemann2013diverse}
J.~Thiemann, N.~Ito, and E.~Vincent, ``The diverse environments multi-channel
  acoustic noise database (demand): A database of multichannel environmental
  noise recordings,'' in \emph{Proceedings of Meetings on Acoustics ICA2013},
  vol.~19, no.~1.\hskip 1em plus 0.5em minus 0.4em\relax Acoustical Society of
  America, 2013, p. 035081.

\bibitem{warden2018speech}
P.~Warden, ``Speech commands: A dataset for limited-vocabulary speech
  recognition,'' \emph{arXiv preprint arXiv:1804.03209}, 2018.

\end{thebibliography}

\end{document}